\documentclass[lettersize,journal]{IEEEtran}
\usepackage{amsmath,amsfonts}
\usepackage{subfigure}
\usepackage{algorithmic}
\usepackage{algorithm}
\usepackage{array}
\usepackage[caption=false,font=normalsize,labelfont=sf,textfont=sf]{subfig}
\usepackage{textcomp}
\usepackage{stfloats}
\usepackage{url}
\usepackage{verbatim}
\usepackage{graphicx}
\usepackage{cite}
\usepackage{multirow}
\usepackage{float}

\begin{document}

\title{Asymmetric Diffusion Based Channel-Adaptive Secure Wireless Semantic Communications}

\author{Xintian Ren,~\IEEEmembership{Student Member, IEEE,} Jun Wu,~\IEEEmembership{Senior Member, IEEE,} Hansong Xu,~\IEEEmembership{Member, IEEE,} \\ Qianqian Pan,~\IEEEmembership{Member, IEEE}
\thanks{
This work was supported in part by the JSPS KAKENHI under Grants 23K11072, in part by the China Scholarship Council Program, and in part by the National Natural Science Foundation of China under Grants U21B2019 and 61972255.
}
\thanks{
Xintian Ren is with the School of Electronic Information and Electrical Engineering, Shanghai Jiao Tong University, Shanghai, 200240, China. Email: xintian\_ren@sjtu.edu.cn.
}
\thanks{Jun Wu is with the Graduate School of Information, Production and Systems, Waseda University, Fukuoka, 808-0135, Japan. Email: junwu@aoni.waseda.jp.}
\thanks{Hansong Xu is with the School of Electronic Information and Electrical Engineering and the Collaborative Innovation Center of Shanghai Industrial Internet, Shanghai Jiao Tong University, Shanghai, 200240, China. Email: hansongxu@sjtu.edu.cn.}
\thanks{Qianqian Pan is with the Graduate School of Engineering, The University of Tokyo, Tokyo, 113-0033, Japan. Email: panqianqian@g.ecc.u-tokyo.ac.jp.}
}

\markboth{Journal of \LaTeX\ Class Files,~Vol.~14, No.~8, August~2021}%
{Shell \MakeLowercase{\textit{et al.}}: A Sample Article Using IEEEtran.cls for IEEE Journals}

\maketitle

\begin{abstract}

Semantic communication has emerged as a new deep learning-based communication paradigm that drives the research of end-to-end data transmission in tasks like image classification, and image reconstruction.
However, the security problem caused by semantic attacks has not been well explored, resulting in vulnerabilities within semantic communication systems exposed to potential semantic perturbations.
In this paper, we propose a secure semantic communication system, DiffuSeC, which leverages the diffusion model and deep reinforcement learning (DRL) to address this issue.
With the diffusing module in the sender end and the asymmetric denoising module in the receiver end, the DiffuSeC mitigates the perturbations added by semantic attacks, including data source attacks and channel attacks.
To further improve the robustness under unstable channel conditions caused by semantic attacks, we developed a DRL-based channel-adaptive diffusion step selection scheme to achieve stable performance under fluctuating environments.
A timestep synchronization scheme is designed for diffusion timestep coordination between the two ends.
Simulation results demonstrate that the proposed DiffuSeC shows higher robust accuracy than previous works under a wide range of channel conditions, and can quickly adjust the model state according to signal-to-noise ratios (SNRs) in unstable environments.

\end{abstract}


\section{Introduction}
As a new communication paradigm, semantic communication, which extracts and transmits task-oriented information, has received increasing attention for its increased robustness and reduction of the transmission burden \cite{iyer2023survey}.
Semantic communication, which takes advantage of Artificial Intelligence (AI) technologies, extracts and only transmits the "meaning" of the data rather than accurate bits\cite{luo2022semantic}.
It has been demonstrated to support various AI-related tasks like machine translation, speech recognition, and visual question answering\cite{weng2023deep, xie2021task}.
Thus, semantic communication is predominantly an AI-driven communication system.

Although lots of attention has been put on semantic communication, the secure semantic communication system in defense of semantic attacks which bring great potential information security risks to semantic communication users, is still in its infancy \cite{du2023rethinking}. 
Semantic attacks, which are considered the attack that causes misunderstanding of semantic information and decoding errors \cite{hu2023robust}, can be produced in the two major resources, including data source attacks and channel attacks.
Unlike other attack methods that aim to simply confuse the bits of the data, the semantic attack focuses on the semantic meaning of the message. 
For example, in a task where the sender sends a picture through semantic communication to the receiver to build a training database for image classification.
The malicious attacker may disturb the signal and secretly alter the semantic meaning from "car" to "horse"\cite{ma2023task}. 
As semantic communication is predicted to become a core paradigm in the sixth-generation (6G) \cite{yang2022semantic}, semantic attacks could cause more serious accidents in the 6G networks\cite{kaewpuang2023cooperative, xi2018sema, pokhrel2022understand}, such as in scenarios like the Industrial Internet of Things (IIoT)\cite{li2022multitentacle, pan2021joint}, vehicular networks\cite{kang2023adversarial, zhang2021sema}, and the Metaverse\cite{li2023secure, wang2023semantic} for their close connection with industrial devices, vehicles, and interactive devices. 

As semantic communication systems are essentially based on AI, the currently most used implementation of semantic attacks is through adversarial attacks on semantic communication models\cite{hu2023robust}. 
Adversarial attacks are a class of attacks on AI models that involve intentionally crafting inputs, such as adding human-invisible noises, that cause the model to produce incorrect outputs. 
In this way, the semantic information for the data can be distorted by the noises without being noticed, and the whole system becomes insecure to the users.
However, the method of defending against semantic attacks has not been widely explored, and as far as we know, all the current works have been conducted according to adversarial training. 
In consideration of the diverse data resources in the wireless network, it's hard to build a semantic communication robust to all the adversarial attacks through adversarial training targeted for some specific attack methods.

Thus, we employ a new adversarial defending method based on input denoising, called diffusion purification, to purify transmitted images back to clean images in the semantic communication system.
Diffusion purification was originally proposed with the booming research of a deep generative model, named diffusion model, which has emerged as a powerful generative model that has record-breaking performance in many applications, such as image synthesis, video generation, and molecule design\cite{yang2022diffusion}. 
It also has shown significant performances in adversarial purification, which purifies attacked images into clean images with a standalone purification model\cite{yoon2021adversarial}.
Although the diffusion model has been employed in semantic communication systems in some prior works to build a system robust to channel noises, there's still negligible paper discussing the availability and method of adopting diffusion purification against semantic attacks and therefore building a secure and robust semantic communication system.

One of the most important factors of the diffusion model is the size of diffusion timesteps, especially in image purification tasks, where the diffusion model limits its diffusion timestep size to a small level to preserve the structure of the original image. 
Some works have conducted DRL to search for the appropriate step size for image purification tasks, while since no discussion has been given under the wireless communication scenario which has extra noises, there's an urgent request for an exploration of a unique diffusion purification scheme for semantic communication systems.

Being inspired by the image purification of the diffusion model, we take a step further and aim to utilize image denoising and recovery in one semantic communication for a secure and robust system.
In this work, we propose a secure semantic communication system with asymmetric diffusion and a DRL-based channel-adaptive diffusion scheme, which shows better security under semantic attacks and higher robustness in fluctuating transmission environments.
To be specific, our contribution can be summarized as follows:
\begin{itemize}
    \item For the semantic attacks from the data source and channel, we propose a secure semantic communication system named DiffuSeC, in which a diffusing module and an asymmetric denoising module are developed for communication scenarios with semantic attacks.
    \item We propose a novel asymmetric diffusion scheme, which has diverse diffusion timesteps in the two ends of the semantic communication, to mitigate the perturbations generated by malicious attackers in the data source and channel. A synchronization scheme is also designed for end-to-end diffusion timestep coordination.
    \item To further raise the system robustness under unstable channel conditions, a channel-adaptive diffusion scheme based on DDPG is employed to quickly adjust the timesteps according to the channel condition while maintaining the image quality and removing the perturbations.
\end{itemize}

\section{Related Works}
The consideration of the possibility of adversarial attacks extending to semantic communication should not be underestimated, as it has the potential to introduce semantic attacks and consequently compromise the security of semantic communication systems.
Adversarial attacks have become an increasingly important area of research in AI for the security concerns they bring in DNN-based critical life applications \cite{khamaiseh2022adversarial}. 
There have been many different adversarial attack methods, such as Fast Gradient Sign Method (FGSM) \cite{goodfellow2015explaining}, Projected Gradient Descent (PGD) \cite{madry2018towards}, Fast Gradient Method (FGM) \cite{miyato2017adversarial}, Deepfool algorithm \cite{moosavi2016deepfool}, Carlini and Wagner Attacks (C\&W) \cite{carlini2017towards}, and researchers are constantly developing new methods and improving existing ones. 
Many of these attack methods have been employed to create semantic attacks in semantic communication systems. 
The authors of \cite{hu2022robust} applied FGSM for creating semantic attacks for images in the semantic communication system.
In \cite{peng2022robust}, the authors utilized FGM to generate semantic attacks that render models to misunderstand text semantics.
Hu \textit{et al.} \cite{hu2023robust} employed iterative FGSM-based and PGD-based methods to generate sample-dependent and sample-independent semantic attacks.

\begin{figure}[t]
    \centering
    \includegraphics[width=\linewidth, scale=0.60]{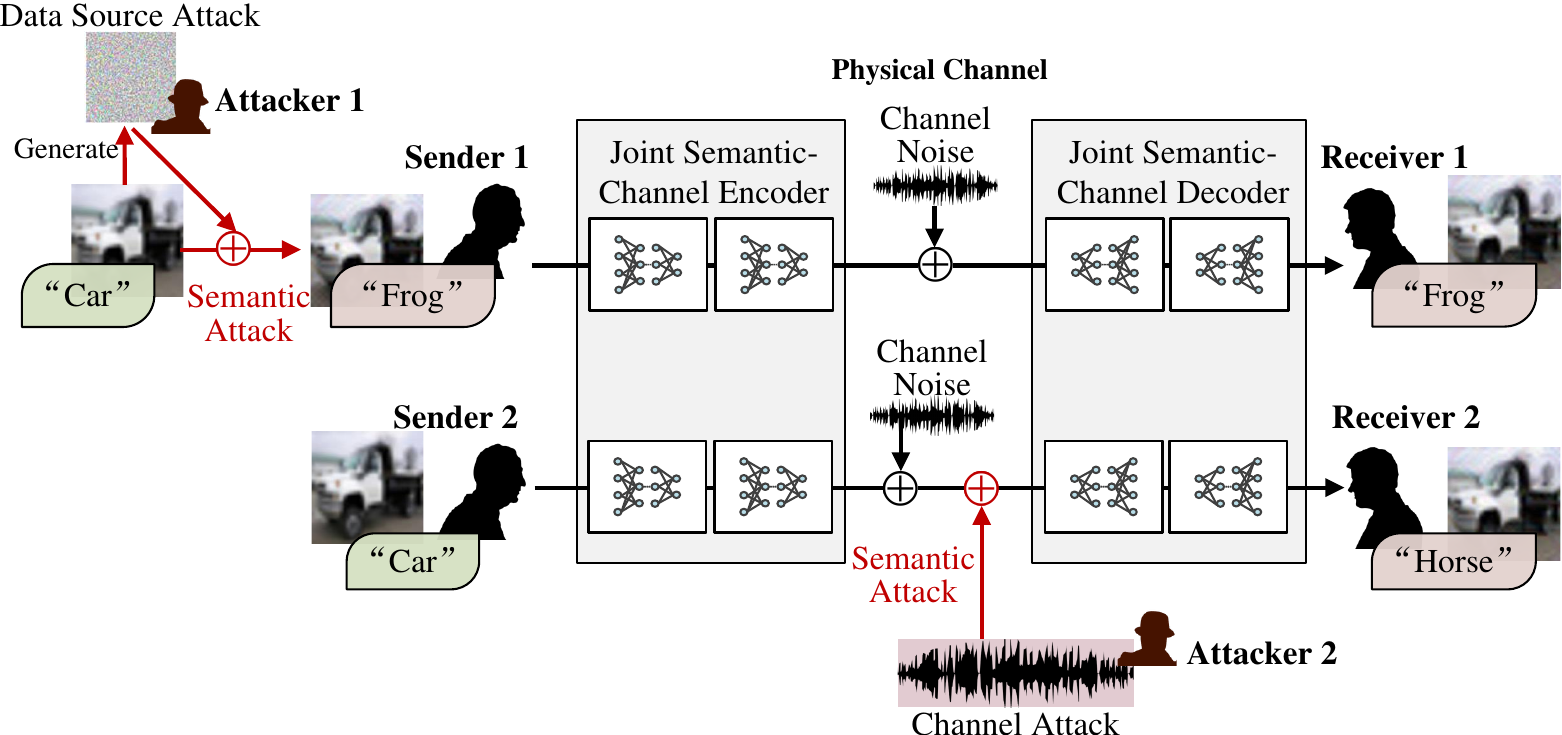}
    \caption{The data source semantic attack and channel semantic attack of semantic communication.}
\end{figure}

To remove the semantic perturbation added by adversarial methods, some efforts have been made within the context of semantic communication systems. 
The authors in \cite{hu2023robust} proposed a masked VQ-VAE-enabled codebook for a robust semantic communication and applied adversarial training to combat the semantic attack, the simulation results showed a significant improvement in robustness against semantic attack.
Peng \textit{et al.} \cite{peng2022robust} applied FGSM to eliminate the inference of the semantic attack in a semantic communication for text.
Nan \textit{et al.} \cite{nan2023physical} focused on physical-layer adversarial attacks, and proposed a physical-layer adversarial perturbation generator that aims to craft semantic adversaries and introduce a novel adversarial training method SemMixed to harden the semantic communication against the attacks.

However, the prior works only employed adversarial training and thus the semantic communication systems can only guarantee their robustness under the specific attacks they are trained for. 
This limited robustness is insufficient for a system with diverse transmission objects and could face any type of semantic attack.
To eliminate perturbations from different types of adversarial attacks, many methods based on adversarial denoising have been developed \cite{akhtar2021advances}, such as the denoising methods in \cite{shi2021online, xie2019feature, gupta2019ciidefence}.
Among them, a variety of strategies based on diffusion models have been proposed to make image classifiers resistant to adversarial attacks.
Diffusion models are deep generative models that are based on two stages, a forward diffusing stage and a reverse denoising stage \cite{croitoru2023diffusion}.
The diffusion model was first proposed in \cite{ho2020denoising}, which presents the Denoising Diffusion Probabilistic Model (DDPM) and offers rigorous mathematical derivations, and refines the inference process. 
The diffusion model overtook the Generative Adversarial Networks (GANs) and reached new state-of-the-art across various domains including robust learning.

While adversarial training is considered a standard defense method, diffusion-based adversarial purification has shown remarkable performance as an alternative defense method that purifies attacked images into clean images \cite{yang2022diffusion}. 
Diffpure \cite{nie2022diffusion} purifies the attacked images by diffusing them with a slight amount of noise and reconstructing the picture with a denoising process. Adaptive Denoising Purification (ADP) \cite{yoon2021adversarial} illustrates that an energy-based model trained with denoising score matching can quickly purify attacked images within a few steps. GDMP \cite{wang2022guided} further equips the diffusion purification model with additional guidance to retain the consistency between the purified images and the original ones.

As one of the deterministic hyperparameters of the effectiveness of diffusion purification, the size of the diffusion and denoising timesteps are discussed in some prior works.
The authors in \cite{nie2022diffusion} have considered the impact of diffusion timestep and compared the accuracy under different timesteps, through the experiments they demonstrate that the choice of timestep is a trade-off between purifying the local perturbations and preserving the global structures.
Yoon \textit{et al.} \cite{yoon2021adversarial} proposed a simple yet effective adaptation scheme with a formula that can choose proper stepsizes during the purification, their evaluation results showed greater robustness than other methods.
Other works gave the robust accuracy under different stepsizes but did not give a specific scheme for the timestep decision \cite{wang2022guided, blau2022threat, wu2022guided, sun2022pointdp}.
To solve the stepsize problem in an unstable new scenario, which is to purify perturbations under wireless semantic communication, we employ a DRL approach to actively adjust the timestep.
Some previous works have shown adaptive semantic communications outperform existing methods\cite{dai2023toward, zhang2023predictive}.
There are various DRL methods like Deep Q-Network (DQN) \cite{mnih2013playing}, Deep Deterministic Policy Gradient (DDPG) \cite{lillicrap2015continuous}, Asynchronous Advantage Actor-Critic (A3C) \cite{mnih2016asynchronous}.
Among them, the DDPG method is selected in this paper to solve the problem of stepsize choosing because its continuous action space fits the large range of timestep.

\begin{figure*}[t]
    \centering
    \includegraphics[width=\linewidth, scale=0.60]{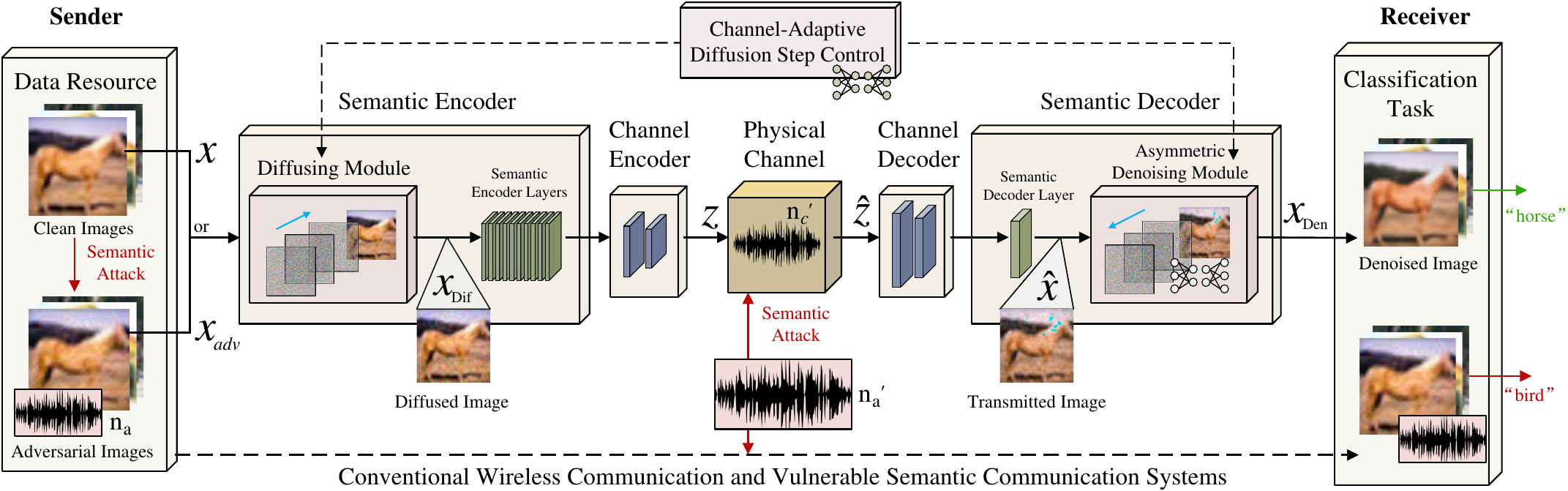}
    \caption{The system architecture of the proposed secure semantic communication.}
\end{figure*}

\section{Secure Semantic Communication System Architecture}
In this section, we give the definition of semantic attacks and propose the system model of the secure semantic communication system.

\subsection{Semantic Attacks}
As shown in Fig. 1, the semantic attacks have two derivations: adversarial perturbations from data source attacks and signal disturbs from channel attacks, which are respectively generated before the transmission and during the transmission. 

Data source attacks mean that the information sender's image collection is polluted by attackers and adversarial images are injected into the database or image stores. 
In consideration of the similarity between the original images and maliciously modified images, this type of attack is harder to find by human users. 
For an adversarial image $x_{adv}$ and its original image $x$, the relation between them can be shown as
$x_{adv}=x+n_a$,
where $n_a$ is the invisible adversarial perturbation in $x_{adv}$ that modifies the classification result of $x$.

The channel attack adds more noise to the transmission signal except the natural noises that are inevitable because of the natural feature of the wireless communication process.
The data $x'$ that goes through the physical channel can be expressed as
${x}' = \mathcal {H}({x}) + n_c$,
where $x$ is the sent data, $\mathcal {H}(\cdot)$ is the channel matrix, and $n_c$ is the channel noise.
To simplify the model, in this work we mainly consider the Additive White Gaussian Noise (AWGN) in the physical channel, so the formula can be simplified as
\begin{equation}
    {x}' = \mathcal {H}({x}) + n_c \approx {x} + {n_c}',
\end{equation}
where ${n_c}'$ represents all noises after data $x$ goes through the physical channel.
The channel attack makes the channel condition unstable so as to mess up or disturb the transmission of data $x$, here we assume the semantic perturbation is also AWGN, the attack can then be written as 
\begin{equation}
    {x}' = \mathcal {H}(x) + n_c + {n_a}' \approx x + {n_c}' + {n_a}',
\end{equation}
where $x'$ is the received data, and ${n_a}'$ is the malicious noise that disturbs the wireless signal.

\subsection{Secure Semantic Communication System Model}

The secure semantic communication we propose has some typical components of regular semantic communication, including the joint semantic-channel encoder and decoder. 
We also adopt a diffusing module and an asymmetric denoising module to enhance the robustness against semantic attacks mentioned above. The architecture of the proposed system is shown in Fig. 2.

Firstly, the images go through the diffusing module which dominates the adversarial perturbations by gradually adding Gaussian noises.
The negligible Gaussian noises are added to the source images $\boldsymbol{x} \in \mathbb{R}^{B\times H\times W\times 3}$ multiple times, which can be modeled as
\begin{equation} 
\boldsymbol{x}_{Dif} = D(\boldsymbol{x}; t_{D}, \sigma_D),
\end{equation} 
where $\boldsymbol{x}_{Dif}\in \mathbb{R}^{B\times H\times W\times 3}$ is the images produced by the diffusing module, $B$ is the batch size, $H$ and $W$ stands for the height and width of the pictures respectively, $D(\cdot; t_{D}, \sigma_D)$ is the process of adding noise to every image in $\boldsymbol{x}$, $t_{D} \in \mathbb{C}$ is the number of diffusing timesteps, and $\sigma_D$ is the parameter to control the amount of Gaussian noise added every timestep.

Then the images go through the joint semantic-channel encoder, including neural networks in the semantic encoder and channel encoder, to be converted into semantic information for transmission, which can be expressed as
\begin{equation} 
\begin{split}
\boldsymbol{z} &= C(S(\boldsymbol{x}_{Dif}; \boldsymbol{\sigma}_S) ; \boldsymbol{\sigma}_C) \\
&= F(\boldsymbol{x}_{Dif}; \boldsymbol{\sigma}_S, \boldsymbol{\sigma}_C),\\
\end{split}
\end{equation} 
where $\boldsymbol{z} \in \mathbb{R}^{B\times L}$ is the extracted semantic information of images $\boldsymbol{x}_{Dif}$, $L$ denotes the length of semantic information of every image, $S(\cdot; \boldsymbol{\sigma}_S)$ is the semantic encoder model with parameters $\boldsymbol{\sigma}_S$, and $C(\cdot; \boldsymbol{\sigma}_C)$ is the channel encoder model with parameters $\boldsymbol{\sigma}_C$. 

The extracted information is subsequently sent by physical channels, which have natural physical noises, according to formula (1), the information received by the receiver can be simplified as
\begin{equation} 
\hat{\boldsymbol{z}} = \mathcal {H}(\boldsymbol{z}) + n_c \approx \boldsymbol{z} + {n_c}',
\end{equation} 
where $\hat{\boldsymbol{z}} \in \mathbb{R}^{B\times L}$ denotes the semantic information received, ${n_c}' \in \mathbb{R}^{B\times L}$. 

The joint semantic-channel decoder then recovers the semantic information back to image mode by
\begin{equation} 
\begin{split}
\hat{\boldsymbol{x}} &= C^{-1}(S^{-1}(\hat{\boldsymbol{z}}; {\boldsymbol{\sigma}_S}') ; {\boldsymbol{\sigma}_C}') \\
&= F^{-1}(\hat{\boldsymbol{z}}; {\boldsymbol{\sigma}_S}' ,{\boldsymbol{\sigma}_C}'),\\
\end{split}
\end{equation}
where $\hat{\boldsymbol{x}}\in \mathbb{R}^{B\times H\times W\times 3}$ is the image decoded from the semantic information, $S^{-1}(\cdot; {\boldsymbol{\sigma}_S}')$ is the semantic decoder model with parameters ${\boldsymbol{\sigma}_S}'$, and $C^{-1}(\cdot; {\boldsymbol{\sigma}_C}')$ is the channel decoder model with parameters ${\boldsymbol{\sigma}_C}'$.

Finally, based on the transmitted images, the asymmetric semantic denoising module generates purified images by gradually shedding off noises from the images. The denoising process can be expressed as
\begin{equation} 
\boldsymbol{x}_{Den} = P(\hat{\boldsymbol{x}}; t_{P}, \boldsymbol{\delta}_P),
\end{equation}
where $\boldsymbol{x}_{Den}\in \mathbb{R}^{B\times H\times W\times 3}$ is the images cleaned from semantic attacks, $P(\cdot;t_{P},\boldsymbol{\delta}_P)$ is the denoising model with parameters $\boldsymbol{\delta}_P$ which gradually shed off the semantic perturbations from the images in $t_{P}$ timesteps.

Thus, the problem of mitigating the semantic attacks in the proposed secure semantic communication system can be formulated as follows:
\begin{equation}
\begin{split}
    \min &~ L(\boldsymbol{x},\boldsymbol{x}_{Den})\\ & = L(\boldsymbol{x},P(F^{-1}(F(D(\boldsymbol{x}+\boldsymbol{n}_a))+{\boldsymbol{n}_c}'+{\boldsymbol{n}_a}'))),
    \end{split}
\end{equation}
where $L$ denotes the un-cleaned semantic perturbations in the images $\boldsymbol{x}_{Den}$ compared to the original images $\boldsymbol{x}$, $\boldsymbol{n}_a, {\boldsymbol{n}_c}',$ and ${\boldsymbol{n}_a}'$ are respectively the adversarial perturbations added to the images before the transmission, the noises in the physical channel, and the perturbations added during the transmission.

\section{Asymmetric Channel-adaptive Diffusion for Semantic Attack Mitigation}
In this work, we divide the two DDPM processes into two ends: the diffusing process, which is adopted before the transmission by the sender, and the asymmetric denoising process, which is handled at the receiver. 
Due to the communication process, the two modules are not as symmetric as the normal diffusion model, so we give a demonstration of the asymmetry and design a timestep synchronization scheme.
As the key to eliminating the semantic attacks while reserving the image features, the DRL-based channel-adaptive diffusion step selection scheme is then introduced in this section.

\subsection{Diffusing Process}
The diffusing process has one Markov process, which gradually adds noise to the image waiting to be sent. 
In the original DDPM process, noises are added to clean images until they are in Gaussian distribution, which can be shown as
\begin{equation} 
q(x^1, ..., x^T| x^0) = \prod_{t=1}^{T} q(x^t|x^{t-1}),  
\end{equation} 
where $T$ is the total timesteps it takes to add noises until a clean image becomes a Gaussian-distributed image, and
\begin{equation} 
q(x^t|x^{t-1})=\mathcal{N}(x^t; \sqrt{1-\beta_t}x^{t-1},\beta_t \boldsymbol{I})
\end{equation} 
where $\beta_t$’s are predefined small positive constants. Consider $\alpha_t=1-\beta_t, \hat{\alpha}_t= \prod_{i=1}^t \alpha_i$, then 
\begin{equation} 
q(x^t|x^0)=\mathcal{N}(x^t; \sqrt{\hat{\alpha}_t}x^{0},(1-\alpha_t)\boldsymbol{I}).
\end{equation} 
It is obvious that $x^t$ can be directly obtained from the following equation,
\begin{equation} 
x^t=\sqrt{\hat{\alpha}_t}x^0 + \sqrt{1-\hat{\alpha}_t}\boldsymbol{n},
\end{equation} 
where $\boldsymbol{n}$ is a standard Gaussian noise.

As for image purification, this process can be utilized to destroy and remove semantic perturbations in an image, including perturbations from data source attacks and channel attacks. In the diffusing process, we assume that $t^D$ timesteps will be taken to purify the images, then after the adversarial attack on the sender side, the images processed by the diffusing module can be expressed as
\begin{equation} 
\begin{split}
    \boldsymbol{x}^{t_D} & = \sqrt{\hat{\alpha}_{t_D}}\boldsymbol{x}_{adv} + \sqrt{1-\hat{\alpha}_{t_D}}\boldsymbol{n}, \\
    & = \sqrt{\hat{\alpha}_{t_D}} \boldsymbol{x}+ \sqrt{\hat{\alpha}_{t_D}}\boldsymbol{n}_a + \sqrt{1-\hat{\alpha}_{t_D}}\boldsymbol{n}.
\end{split}
\end{equation}

As $t_D$ increases, the $\sqrt{\hat{\alpha}_{t_D}}$ gets smaller, and $1-\hat{\alpha}_{t_D}$ gets bigger.
It is widely acknowledged that $\boldsymbol{n}_a$ is relatively small compared to $\boldsymbol{x}$, so when $1-\sqrt{\hat{\alpha}_{t_D}}$ gets big enough, the Gaussian noise we add $\boldsymbol{n}$ is possible to merge the adversarial noises, and be eliminated in the following denoising timesteps.

\subsection{Asymmetric Denoising Process}
After the semantic attacks before and during the transmission, and going through the physical channel, the images now can be regarded as  
\begin{align}
    \hat{\boldsymbol{x}} & = F^{-1}(F(D(\boldsymbol{x}+\boldsymbol{n}_a))+{\boldsymbol{n}_c}'+{\boldsymbol{n}_a}')\\
    &=F^{-1}(F(\sqrt{\hat{\alpha}_{t_D}} (\boldsymbol{x}+\boldsymbol{n}_a)+\sqrt{1-\hat{\alpha}_{t_D}} \boldsymbol{n})+{\boldsymbol{n}_c}'+{\boldsymbol{n}_a}')\\
    & = \epsilon (\sqrt{\hat{\alpha}_{t_D}} (\boldsymbol{x}+\boldsymbol{n}_a) +\sqrt{1-\hat{\alpha}_{t_D}} \boldsymbol{n})+ F^{-1}({\boldsymbol{n}_c}'+{\boldsymbol{n}_a}').
\end{align}

According to the goal of semantic communication, the joint semantic-channel encoder and decoder are trained to extract the semantic information and recover the image on the other side, so the images are not scaled after the transmission process. Thus, the parameter $\epsilon$ is expected to approximate 1, then the formula can be rewritten as
\begin{equation}
    \hat{\boldsymbol{x}} \approx \sqrt{\hat{\alpha}_{t_D}} (\boldsymbol{x} + \boldsymbol{n}_a) + \sqrt{1-\hat{\alpha}_{t_D}} \boldsymbol{n} + F^{-1}({\boldsymbol{n}_c}'+{\boldsymbol{n}_a}')\\
\end{equation} 
\begin{equation}
    \begin{split}
      &= \sqrt{\hat{\alpha}_{t_D}} \boldsymbol{x} + \sqrt{\hat{\alpha}_{t_D}}(\boldsymbol{n}_a + \dfrac{1}{\sqrt{\hat{\alpha}_{t_D}}}F^{-1}({\boldsymbol{n}_c}'+{\boldsymbol{n}_a}')) \\
    & ~~~+ \sqrt{1-\hat{\alpha}_{t_D}} \boldsymbol{n}.
\end{split}
\end{equation}

For the adversarial semantic perturbations $\boldsymbol{n}_a$, when the $\sqrt{1-\hat{\alpha}_{t_D}} $ gets big enough, the noises $\boldsymbol{n}$ can merge the noises $\boldsymbol{n}_a$. And for two other noises, ${\boldsymbol{n}_c}'$ and ${\boldsymbol{n}_a}'$ represent the condition of the physical channel. When ${\boldsymbol{n}_c}'$ and ${\boldsymbol{n}_a}'$ are small enough, the $\boldsymbol{n}$ is able to merge them as the way it merges $\boldsymbol{n}_a$. Although the ${\boldsymbol{n}_c}'$ or ${\boldsymbol{n}_a}'$ can grow to an extent that won't be merged by $\boldsymbol{n}$, this module still raises the robustness of the semantic communication system.
In this way, the adversarial semantic perturbations $\boldsymbol{n}_a$ in the image set $\hat{\boldsymbol{x}} $ can be removed by the denoising process, and the semantic communication system becomes more robust to the signal disturbances ${\boldsymbol{n}_c}'$ and ${\boldsymbol{n}_a}'$.

The denoising process of DDPM is a Markov process that predicts and eliminates the noise that is added in the diffusing process. The denoising process can be defined as
\begin{equation} 
p_{\boldsymbol{\delta}}(x^0, ..., x^{T-1}|x^T)=\prod_{t=1}^{T} p_{\boldsymbol{\delta}}(x^{t-1}|x^{t}),
\end{equation} 
where 
\begin{equation} 
p_{\boldsymbol{\delta}}(x^{t-1}|x^{t})=\mathcal{N}(x^{t-1}; \mu_{\boldsymbol{\delta}}(x^t,t), \sigma_t^2 \boldsymbol{I}).
\end{equation} 
where the mean $\mu_{\boldsymbol{\delta}}(x^t,t)$ is a neural network with parameters $\boldsymbol{\delta}$, and $\sigma_t^2$ is constants related to timestep.

Instead of adopting the denoising process on the sender side, this process is delayed in our work to purify images after the transmission. 
Setting in the receiver allows this process to not only eliminate the data source attacks but also deal with the natural noises in the physical channel and the channel attacks.

The denoising process for image purification can be expressed as
\begin{equation} 
p_{\boldsymbol{\delta}}(x^0, ..., x^{t_P -1}|x^{t_P})=\prod_{t=1}^{t_P} p_{\boldsymbol{\delta}}(x^{t-1}|x^{t}),
\end{equation} 
where $t_P$ is the size of the denoise timestep of image $x$, and $x^{0}$ is the final image produced by our semantic communication system, namely $x_{final}$.

\begin{figure}[t]
    \centering
    \includegraphics[width=\linewidth, scale=0.60]{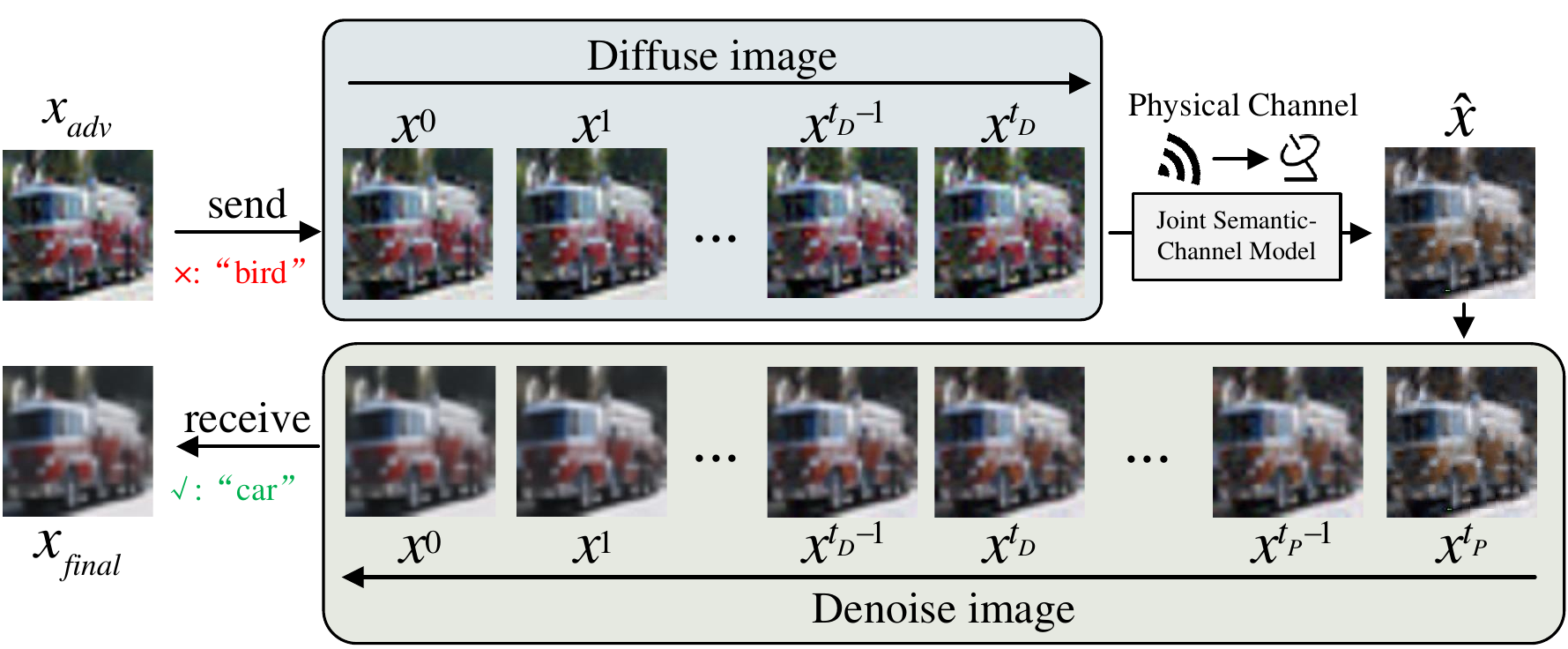}
    \caption{An illustration of the diffusing process and asymmetric denoising process mitigating semantic attack.}
\end{figure}

In the conventional diffusion model, the steps size of diffusing and reversing should be the same, as we showed in formulas (9) and (19). 
This balance is still kept in the former works of diffusion purification, in which the steps taken to add noises and purify the images remained the same.
However, as we illustrate in Fig. 3, the wireless communication between the diffusing and denoising employs more noise and breaks the balance between these two processes.
To be specific, as shown in formula (13), with the proposed system, the input image $\boldsymbol{x}_{adv}$ is added with relatively small Gaussian noises, and the channel noises that are not mitigated by the joint semantic-channel model. Instead of being directly sent into the denoising process, the images are equipped with more noise
\begin{equation}
    \begin{split}
    \boldsymbol{n}_{plus} & = \hat{\boldsymbol{x}}-\boldsymbol{x}^{t_D}\\
    & = F^{-1}({\boldsymbol{n}_c}'+{\boldsymbol{n}_a}'),
\end{split}
\end{equation}
where $\boldsymbol{n}_{plus}$ is the noise that causes the asymmetry between the two processes. We aim to reduce the impact of $\boldsymbol{n}_{plus}$ by adding the size of the denoising step from $t_D$ to $t_P$, which can be controlled dynamically by the module in real time.
Here we call the gap between $t_D$ and $t_P$ as plus steps $t_{plus}$, then the formula (21) can be reformed as
\begin{equation}
\begin{split}
 p_{\boldsymbol{\delta}}(x^0, ..., x^{t_P -1}|x^{t_P})= &
 \begin{matrix}
\underbrace {\prod_{t=1}^{t_D} p_{\boldsymbol{\delta}}(x^{t-1}|x^{t}) }\\
\textrm{steps of diffusing}
\end{matrix}
+ 
\begin{matrix}
\underbrace {\prod_{t=t_D}^{t_P} p_{\boldsymbol{\delta}}(x^{t-1}|x^{t})},\\
\textrm{plus steps}
\end{matrix}
\\
&t_D < t_P.
\end{split}
\end{equation}

\subsection{Diffusion Timestep Synchronization}
To ensure the proceeding of the diffusing and asymmetric denoise process, the timesteps shared by these two processes should be synchronized in real time. The timestep of adding the noises in the sender should be the same when the noises are eliminated in the receiver. 
Thus, a timestep synchronization process should be taken before and during the transmission of semantic information.

To be specific, a short communication process is implemented before starting a semantic communication process. 
The sender first sends out a short signal to the receiver to test the channel condition between the sender and the receiver.
After receiving the signal, the SNR is measured by the receiver and is leveraged in section IV (D) to select proper timesteps for both the diffusing module and the denoising module.
The diffusing timestep defined by the receiver is then sent back to the original sender for the following semantic communication process.

The whole process of timestep synchronization is short and efficient, which only takes one round of transmission, and the messages are both very brief.
The timestep is synchronized at certain intervals to control the workload of two ends.

\subsection{DRL-Based Channel-Adaptive Diffusion Step Selection}
As demonstrated above, the size of the timesteps is the key to eliminating the semantic attacks. To be specific, it is a trade-off between removing all semantic perturbations but getting a completely different recovered image and insufficient denoising which leaves too much malicious perturbation. We formulate the problem of formula (23) as a Markov Decision Process (MDP) problem. 

The MDP has four components, namely $\{S, A, P, R\}$, $S$ stands for the state space, $A$ represents action space, $P$ is the state transition probability, and $R$ denotes reward. For each cycle, the agent has an observation of the state $s^i \in S$ and chooses an action $a^i \in A$ according to $s^i$. After the action is taken, the agent goes to a new state $s^{i+1}$. Subsequently, a reward $r^i=R(s^i, s^{i+1})$ is fed back to the agent by the environment. We give a detailed statement of the components in the following paragraphs.

\textbf{State space:} The state space of this problem is composed of 3 factors, the timestep size of diffusing $s_{t_D} $, the plus step size $s_{t_{plus}} $, and the SNR of the channel $s_{SNR}$. 
The first two states $s_{t_D}$ and $s_{t_{plus}}$ reflect the state of the semantic communication system, while the last state $s_{SNR}$ denotes the transmission condition of the physical channel. 
According to the former works\cite{nie2022diffusion, yoon2021adversarial, wang2022guided}, the timestep of diffusion purification is mainly in a relatively low range to keep the features of the original images, so that the images after the reverse process share high similarity to the original ones.
So instead of set the two factors as $t_D \in [0, T]$ and $t_{plus} \in [0, T-t_D]$, we set $t_D$ in a range of $[1,t_{D_{max}}]$ and $t_{plus}$ in a range of $[0,t_{{plus}_{max}}]$, following the rules of 
\begin{align}
    & t_{D_{max}}+t_{{plus}_{max}} < T, \\
     \textrm{and~~~}  &~ t_D+t_{plus} < t_{D_{max}}.
\end{align}
As for the channel state, we consider both the unfriendly and good channel conditions and set SNR in a wide range.

\textbf{Action space:} The action space has two components, the modification of diffusing timestep size $a_{t_D} \in [-t_{D_{max}}', t_{D_{max}}'], t_{D_{max}}' \in [0, t_{D_{max}}]$ and plus step size $a_{t_{plus}}  \in [-t_{{plus}_{max}}', t_{{plus}_{max}}'],  t_{{plus}_{max}}' \in [0, t_{{plus}_{max}}]$. 

\textbf{State transition probability:} The state transition probability is defined as  $P_a(s',s)=Pr(s^{i+1}=s'| s^i=s, a^i=a)$. $P_a(s',s)$ represents the probability distribution of $s'$ based on the given state $s$ and chosen action $a$. In this work, the transitions are decided and done in the simulator of our proposed semantic communication environment.

\textbf{Reward function:} The reward function provides the immediate reward $R_a(s,s')$ for transmitting from state $s$ to state $s'$ with action $a$. 
In this paper, we aim to seek a trade-off between purifying all the adversarial semantic perturbations and maintaining the original semantic features in the images. Hence, the reward is defined by three elements: the Structural Similarity Index Measure (SSIM) score of the recovered images, the rate of adversarial images that are not purified, and the rate of images that are purified into a new category yet still mistakenly classified.

The SSIM score denotes the similarity between the original image and the recovered image, the average SSIM score in a batch can be defined as
\begin{equation}
    \text{SSIM}_{avg}(\boldsymbol{x},\boldsymbol{x}_{final})= \frac{1}{B} \sum_{k=0}^{B} \text{SSIM}(x^k, x^k_{final}),
\end{equation}
where $B$ is the batch size of images in one round.

We consider the perception of adversarial images that are not purified by our secure semantic communication system, which indicates the ability of our proposed system to protect users from adversarial attacks. This component is given by
\begin{equation}
\text{Adv}(\boldsymbol{x},\boldsymbol{x}_{adv},\boldsymbol{x}_{final})= \sqrt{\frac{1}{B} \sum_{k=0}^{B} G({x^k},{x}^k_{adv},{x}^k_{final})},
\end{equation}
where 
\begin{equation}
G({x^k},{x}^k_{adv},{x}^k_{final}) = 
\begin{cases}
    1, &\mbox{if } H(x^k) \ne  H({x}^k_{adv}) \mbox{ and } \\ & H({x}^k_{adv}) = H({x}^k_{final}),\\
    0, & \mbox{others},
\end{cases}
\end{equation}
where $H(x)$ is the classification result of image $x$.

Furthermore, to prevent the system from adding too much noise and destroying the original features of images, we set $\text{Err}(\boldsymbol{x},\boldsymbol{x}_{adv},\boldsymbol{x}_{final})$ as the proportion of images that are purified into new but wrong categories, which is denoted by
\begin{equation}
\text{Err}(\boldsymbol{x},\boldsymbol{x}_{adv},\boldsymbol{x}_{final}) = \sqrt{\frac{1}{B} \sum_{k=0}^{B} J({x^k},{x}^k_{adv},{x}^k_{final})},
\end{equation}
where
\begin{equation}
J({x^k},{x}^k_{adv},{x}^k_{final}) = 
\begin{cases}
    1, &\mbox{if } \\& H(x^k) \ne  H({x}^k_{adv}) \ne H({x}^k_{final}),\\
    0, & \mbox{others}.
\end{cases}
\end{equation}

Subsequently, the reward function in time $i$ can be expressed as
\begin{equation}
\begin{split}
     & R^i= \eta(\eta_1(1-\text{SSIM}(\boldsymbol{x}^i, \boldsymbol{x}^i_{final})) +   \\
     & \eta_2(\text{Adv}(\boldsymbol{x}^i,\boldsymbol{x}^i_{adv},\boldsymbol{x}^i_{final})) + \eta_3(\text{Err}(\boldsymbol{x}^i,\boldsymbol{x}^i_{adv},\boldsymbol{x}^i_{final}))),
\end{split}
\end{equation}
where $\eta$ is the factor to limit the reward value, $\eta_1, \eta_2,$ and $\eta_3$ are the parameters to control the contribution of each component in the total reward.

\section{Training Algorithms for Secure Semantic Communication Model}
We divide the training method into two training phases, train ViT-based joint semantic-channel encoder and decoder model, train diffusion purification model, and train channel-adaptive diffusion step selection DDPG model.

\subsection{ViT-based Joint Semantic-Channel Encoder and Decoder Model}
As semantic communication is to minimize the loss in the transmitting process for a specific task, this phase jointly trains the semantic model and the Joint Source-Channel (JSC) model to reduce the loss of image reconstruction. In this phase, the training is divided into three parts to accelerate the training process, including \emph{semantic model training}, \emph{JSC model training}, and \emph{joint semantic-channel training}. 

Firstly, in the \emph{semantic model training}, a semantic model is trained to extract the semantic information and decode them back to images that are as similar as possible to the original images. 
The semantic encoder is a ViT model that encodes the images into embeddings with semantics. 
ViT model has the encoder of the Transformer, which can capture the major semantics in the image through the attention mechanism\cite{vaswani2017attention}. 
For an image reconstruction task, we want the semantic decoder to reconstruct the images from the embeddings as similar to the original image as possible, so we adopted SSIM for the loss function, rather than Mean Squared Error (MSE) or Peak Signal-to-Noise Ratio (PSNR).  
SSIM is a method for quantifying image quality degradation after image compression or data transmission. Unlike other techniques such as MSE or PSNR which estimate absolute errors between two pictures, SSIM focuses more on the perceptual difference.
The SSIM between two images $a$ and $b$ is
\begin{equation}
    \text{SSIM}(a,b)=\frac{(2\mu_{a}\mu_{b}+c_1)(2\sigma_{ab}+c_2)}{(\mu_{a}^2+\mu_{b}^2+c_1)(\sigma_{a}^2+\sigma_{b}^2+c_2)},
\end{equation}
where $\mu_{a}, \mu_{b}$ are the pixel sample mean of images a and b, $\sigma_{a}^2, \sigma_{b}^2$ denote the variance of a and b, $\sigma_{ab}$ is the cross-correlation of the two images, $c_1$ and $c_2$ are two variables to stabilize the division with weak denominator. SSIM is in a range of $[0, 1]$, with 1 denoting that $b$ is completely the same as $a$, and 0 shows that $b$ is completely different.

Thus the loss function can be expressed as
\begin{equation} 
\mathcal{L}_{S}(\boldsymbol{y},\hat{\boldsymbol{y}})=\iota (1- \frac{1}{B_S} \sum^{B_S}_{i=1} \text{SSIM}(\boldsymbol{y},\hat{\boldsymbol{y}}))
\end{equation}
with $\iota$ representing a parameter that is used to control the influence of the SSIM score on the loss,  $\boldsymbol{y}$, $\hat{\boldsymbol{y}}$ are the images before and after going through the semantic model, and $B_S$ denotes the batch size of training.

Secondly, the \emph{JSC model training} part trains a JSC model, which also has an encoder and a decoder, to transmit the semantic information through the physical channel and to defend the physical noises. Although the noise in the channel can cause some errors in the digits, these errors can be mitigated in the semantic decoding, so the embedded semantic information is less affected and more robust.
The JSC encoder and decoder undertake the transmission of digit data, so the MSE loss function is used,
\begin{equation}
    \mathcal{L}_J(\boldsymbol{y},\hat{\boldsymbol{y}})= \frac{1}{B_J}\sum^{B_J}_{i=1} (\boldsymbol{y}-\hat{\boldsymbol{y}})^2,
\end{equation}
where $B_J$ is the batch size.

Finally, the whole model is trained in \emph{joint semantic-channel training} to capture image features, compress and recover the semantic information, and finally get the preliminary transmitted images. This phase is the combination of the first two parts with the loss function of the first part.

\begin{algorithm}[t]
    \caption{Training Procedure of the DDPG Model for Channel-Adaptive Diffusion Step Selection }
    \begin{algorithmic}[1]
        \STATE Initialize the replay buffer $R$ and mini-batch size $B$
        \STATE Randomly initialize the weight of actor net $\theta_\mu$ and critic net $\theta_Q$
        \STATE Set target parameters equal to main parameters ${\theta}_{\mu'} \leftarrow {\theta}_\mu$, ${\theta}_{Q'} \leftarrow {\theta}_Q$
        \FOR{epoch $e=1$ to $E$}
            \STATE Initialize semantic communication scenario, randomly initialize $s^0$ with the observation of SNR in the physical channel
            \FOR{episode $l=1$ to $L$}
                \STATE Get action with action net and behavior noise: $a^l=\mu(s^l|\theta_\mu)+n^l$
                \STATE Perform $a^l$, get the next state $s^{l+1}$ and immediate reward $r^l$ according to equation (31)
                \IF{the replay buffer is not full}
                    \STATE Store the set $(s^l, a^l, r^l,s^{l+1})$ in the buffer $R$
                \ELSE
                    \STATE Replace the oldest set in the $B$ with the set $(s^l, a^l, r^l,s^{l+1})$
                    \STATE Randomly choose $B$ sets to form a mini-batch: $(s^m, a^m, r^m,s^{m+1}), \forall m=1, 2, ..., B$
                    \STATE Calculate the target value $y^m$ by equation (36)
                    \STATE Update the critic online net by minimizing the loss given by (37)
                    \STATE Update the actor online net by the sampled policy gradient given by (38)
                    \STATE Update the target nets:\\
                    $\theta_{Q'}=\tau\theta_{Q}+(1-\tau)\theta_{Q'}$\\
                    $\theta_{\mu'}=\tau\theta_{\mu}+(1-\tau)\theta_{\mu'}$
                \ENDIF
            \ENDFOR
        \ENDFOR
    \end{algorithmic}
\end{algorithm}

\subsection{Diffusion Purification Model in Denoising Module}
We aim to make the diffusion model at the receiver rebuild the native image and remove the impact of semantic attacks at the same time. 
To save computational resources and simplify the setting of the training loss, we concentrate on reconstructing images that are more similar to the clean images.
The input to the model is the images that go through the previous parts of the secure semantic communication system, including being diffused by adding timesteps of noises and being encoded and decoded by the ViT-based joint semantic-channel communication model.

We designed the loss function of the diffusion model as two components:
\begin{equation} 
\begin{split}
&\mathcal{L}_{P}(\boldsymbol{y},{\boldsymbol{y}}') =  \\ 
& \zeta \iota' (1-\frac{1}{B_P}\sum^{B_P}_{i=1}\text{SSIM}(\boldsymbol{y},{\boldsymbol{y}}'))+(1-\zeta) \frac{1}{B_P}\sum^{B_P}_{i=1}(\boldsymbol{y}-{\boldsymbol{y}}')^2
\end{split}
\end{equation}
where $y$ and $y'$ are the clean images and the images go through the whole semantic communication, $B_P$ is the batch size, $\iota'$ is the SSIM influence control parameter, and $\zeta$ is a parameter to adjust the perception of the two components.
As we mentioned before, MSE is a widely used loss function for measuring absolute errors, and SSIM takes image structure as a priority, so we believe the combination of the two loss functions serves better for our task.

\subsection{Channel-Adaptive Diffusion Step Selection DDPG Model}
The problem of deciding the step size of the diffusing and denoising process in an unstable communication environment has large and complex state space and action space. 
The step sizes are continuous integers in a wide range, so we employed DDPG for the adaptive diffusion in the semantic communication system.

We propose a DDPG-based scheme to adaptively control the diffusing and denoising timesteps in different communication environments. The whole training process is based on the semantic communication system that we proposed above.

DDPG is a combination of actor-critic structure and deep neural network (DNN). 
It comprises three elements, actor networks, critic networks, and a replay buffer. 
And for the first two elements, they both have two DNNs inside. 
For the actor networks, we have one DNN called online actor network, $\mu(s;\theta_\mu)$, to choose actions based on the current state $s$, $\theta_\mu$ is the parameters of the network $\mu$; and another DNN named target actor network, $\mu'(s;\theta_{\mu'})$, to record and update the value of $\mu(s;\theta_\mu)$ regularly, where $\theta_{\mu'}$ represents the parameters of network $\mu'$. 
The critic networks also have two DNNs for online and target critic networks, respectively $Q(s, a;\theta_Q)$, $Q'(s, a;\theta_{Q'})$, in which $\theta_Q$ and $\theta_{Q'}$ are the parameters of network $Q$ and $Q'$. 
$Q$ network gives the evaluation of doing action $a$ under state $s$, and the parameters of network $Q'$ is updated periodically by $Q$'s. The replay buffer stores the records as a sequence like $(s^k,a^k,r^k,s^{k+1})$, and provides its storage for future network training.

The algorithm for training the DDPG model is demonstrated in algorithm 1.

Firstly, the replay buffer $R$, mini-batch size $B$, and weight of actor and critic networks are initiated. 
The learning rate of the actor network and the critic network are set as $lr_a$ and $lr_c$. 
A discount factor $\gamma \in [0,1]$ is used to adjust the model's consideration of future rewards, and a soft update factor $\tau$ is employed for the soft update of both actor and critic target networks.

Secondly, for each epoch, we initialize the semantic communication environment. We randomly set the timestep and plus step in the first state $s^0$, and use the observation of the physical channel to initiate the SNR in $s^0$.
We let the episode size be $L$, and for episode $l= 1, 2, ..., L$, we get action $a^l$ with action network evaluating current state $s^l$. 
A small noise $n^l$ is added to $a^l$ for exploring new actions.
Then the next state $s^{l+1}$ is returned by the environment and immediate reward $r^l$ is given by equation (31).

The sequence $(s^l,a^l,r^l,s^{l+1})$ is directly saved to the replay buffer if the buffer is not full, or it will replace the oldest record in the buffer.
A mini-batch is then randomly selected from the buffer for model training, and for $(s^m, a^m, r^m,s^{m+1}), m \in B$, the target value is calculated by
\begin{equation}
    y^m=r^m+\gamma Q'(s^{m+1},\mu'(s^{m+1})|\theta_{Q'}),
\end{equation}
where $y^m$ is the target value in step $m$. The target value $y^m$ is then used to form the loss:
\begin{equation}
    L(\theta_Q)=\frac{1}{B}\sum^m (Q(s^m,a^m|\theta_Q)-y^m)^2.
\end{equation}
By minimizing the $L(\theta_Q)$, we update the critic online network, and the actor online network is updated by the sampled policy gradient
\begin{equation}
    \nabla_{\theta_\mu}J\approx\frac{1}{B}\sum^m \nabla_a Q(s^m,a^m|\theta_Q) \nabla_{\theta_\mu} \mu(s^m|\theta_\mu).
\end{equation}
Finally, we softly update the target networks of both the actor and the critic through their online network.

\section{Simulation Results and Analysis}
In this section, we use numerical results to demonstrate the security under semantic attacks and robustness under fluctuating environments of the proposed secure semantic communication.

\begin{table}[!t]
\centering
\caption{The network architecture of the proposed system.}

\begin{tabular}{|c|c|c|c|}
\hline
 ~ & Name & Units & Activation\\
\hline
\multirow{4}*{Transmitter} & Diffusing & - & - \\
\cline{2-4}
 ~ & Transformer Encoder $\times 10$ & 352(8 heads) & Linear \\
\cline{2-4}
 ~ & Dense & 256 & ReLU\\
\cline{2-4}
 ~ & Dense & 160 & Linear\\
\hline
Channel & AWGN & - & - \\
\hline
\multirow{4}*{Receiver} & Dense & 1024 & ReLU\\
\cline{2-4}
~ & Dense & 352 & Linear\\
\cline{2-4}
~ & Decoder Layer & 300 & Linear\\
\cline{2-4}
~ & Denoising & U-Net \cite{wang2022guided} & - \\
\hline
\end{tabular}
\end{table}

\begin{figure*}[t]
    \subfigure[The rewards of the proposed DiffuSeC.]{
    \begin{minipage}[t]{0.455\linewidth}
        \centering
        \includegraphics[width=\linewidth]{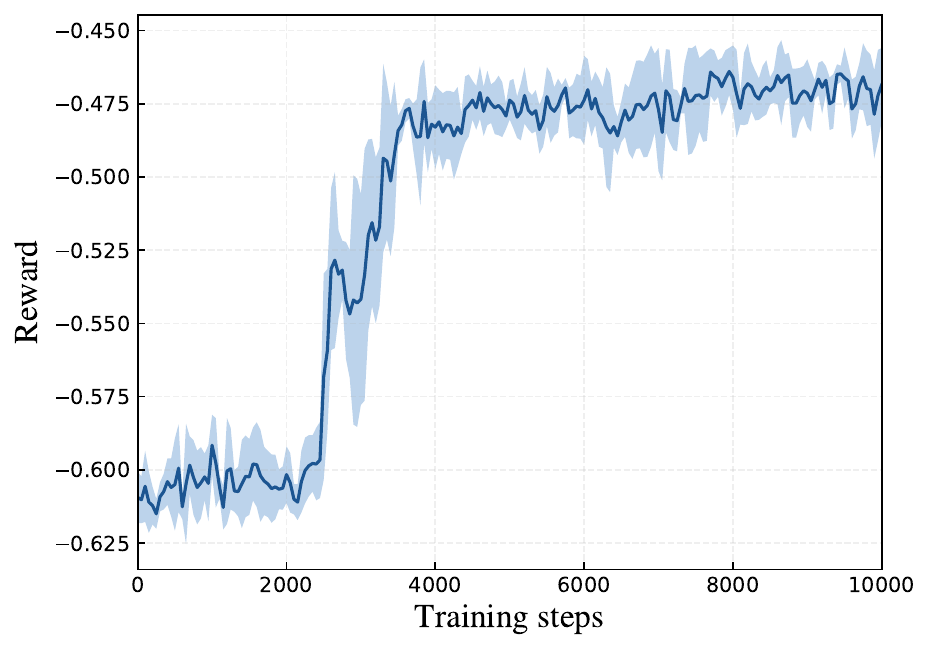}
    \end{minipage}
    }%
    \hspace{0.05cm}
    \subfigure[The SSIM score under different SNRs.]{
    \begin{minipage}[t]{0.44\linewidth}
        \centering
        \includegraphics[width=\linewidth]{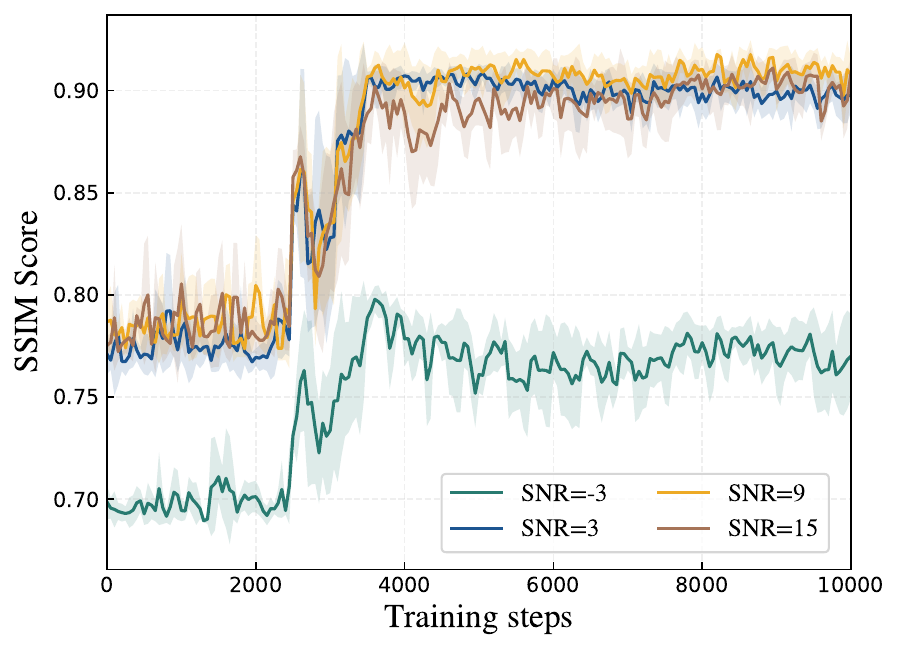}
    \end{minipage}
    }%
    \centering

    \hspace{0.02cm}
    \subfigure[Adversarial error rate under different SNRs.]{
    \begin{minipage}[t]{0.448\linewidth}
        \centering
        \includegraphics[width=\linewidth]{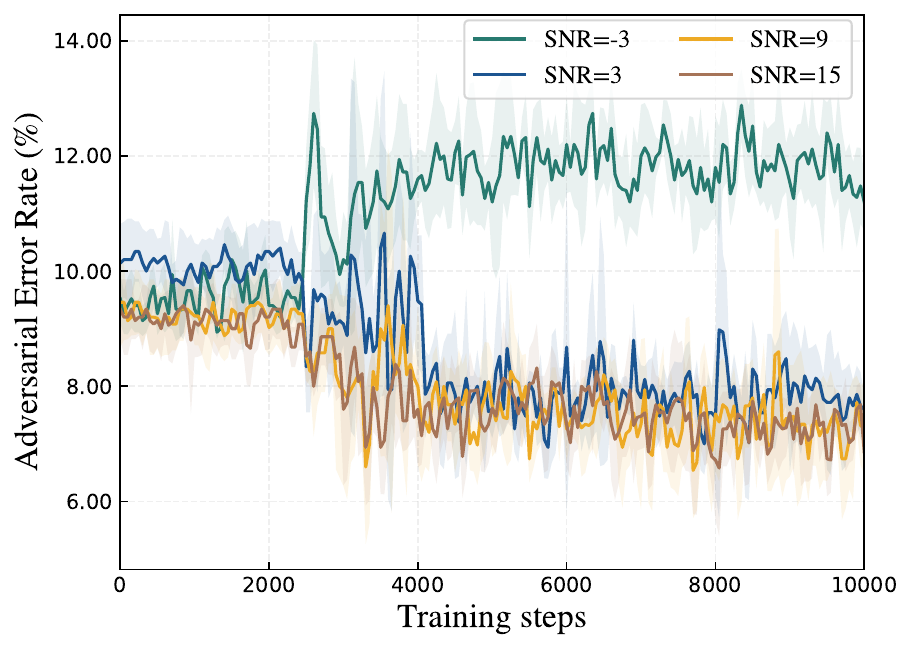}
    \end{minipage}
    }%
    \subfigure[Purification error rate under different SNRs.]{
    \begin{minipage}[t]{0.448\linewidth}
        \centering
        \includegraphics[width=\linewidth]{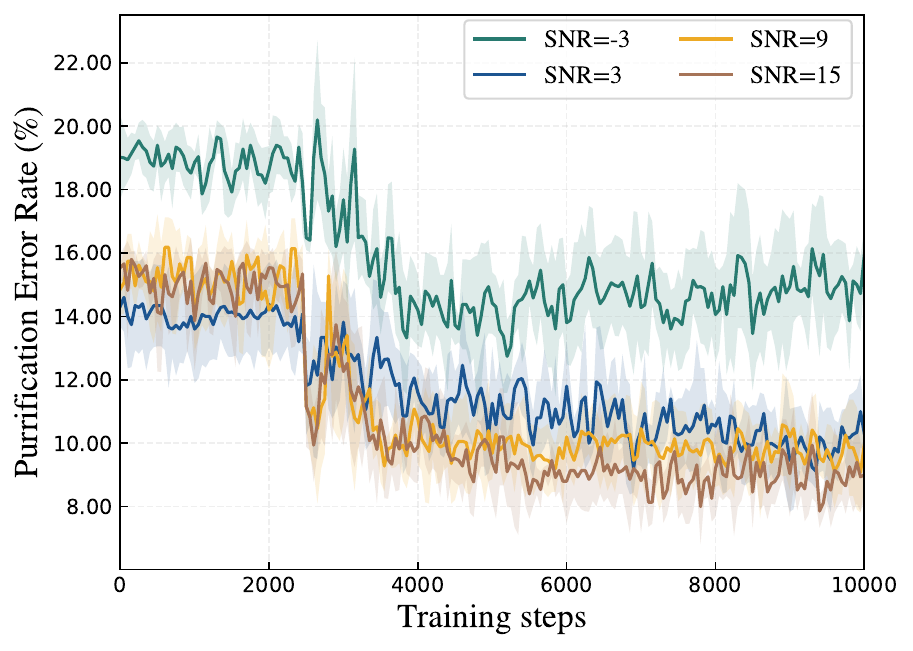}
    \end{minipage}
    }%
    \centering
    \caption{The performance during the training phase in terms of (a) reward, (b) SSIM score, (c) adversarial error rate, and (d) purification error rate.}
\end{figure*}

\subsection{Experimental Settings}
We consider the scenario that images are sent through DiffuSeC for semantic perturbation purification.
For the dataset of this task, we use CIFAR-10, a dataset for image classification that consists of 60, 000 color images in 10 classes, with 6,000 images per class.
The size of the images in the dataset is $32\times 32$. 

The network architecture of the proposed system is presented in Table I. 
The "Diffusing" is the representation of the diffusing process, and the unit of "Denoising" is a U-net model that undertakes the work of the denoising process.
To verify the system's robustness under semantic attacks, we adopt an individual WideResNet model with the architecture of WRN-28-10.

We employed PGD as the adversarial attack method to generate white-box adversarial images in the sender's data resource. The PGD is performed at $\textit{l}_{\infty} \gamma$-ball with $\gamma=8/256$, and the attack step size is 10.

We compare the performance of the following methods:
\begin{itemize}
    \item[$\bullet$] DiffuSeC(proposal): The proposed secure semantic communication with asymmetric diffusion and DRL-based channel-adaptive diffusion scheme.
    \item[$\bullet$] DiffuSeC(plain): The proposed system with symmetric diffusion steps and without channel-adaptive diffusion scheme, the diffusion steps are fixed.
    \item[$\bullet$] Masked VQ-VAE + AT\cite{hu2022robust}: The masked VQ-VAE method with adversarial training.
    \item[$\bullet$] ViT-based: The jointly trained semantic communication composed by conventional ViT, the architecture of semantic and channel encoder and decoder is the same as the DiffuSeC.
    \item[$\bullet$] JPEG + LDPC + AT: The conventional scheme adopts JPEG for the image source coding, Low-Density Parity Check (LDPC) for the channel coding, and the ViT as a classifier with the adversarial training. 
    \item[$\bullet$] JPEG + LDPC: The conventional scheme with JPEG and LDPC.
\end{itemize}

For the parameters in the loss functions in section V, we set $\iota$ and $\iota'$ as 0.5, and $\zeta$ as 0.8. The batch sizes $B_S, B_J$, and $B_P$ are set to 128, 64, and 128.

The settings of the DDPG are as follows: for the state space, $t_{D_{max}}$ and $t_{{plus}_{max}}$ are both 50; for the action space, $t'_{D_{max}}$ and $t'_{{plus}_{max}}$ are 25; for the reward function, we set the batch size $B$ as 256, $\eta, \eta_1, \eta_2, $and $\eta_3$ are respectively $1, -0.8, -0.7,$ and $-0.5$; the replay buffer size is $1\times 10^{6}$, the learning rate of critic network is $1\times 10^{-4}$ and $1\times 10^{-5}$ for the actor network, the discount factor $\gamma$ is 0.99, and the soft update parameter $\tau$ is $5\times 10^{-3}$.

The networks are organized as critic networks with 3 fully connected layers, where the size of the hidden layers is 256, and actor networks with 3 fully connected layers, which also have a hidden layer of 256.
The two former layers of the networks are activated by ReLU, and the final layers of the actor networks are activated by Tanh.

\subsection{Training Phase of Channel-Adaptive Diffusion Scheme}
We let the state be updated 3 times every episode and evaluate the average reward in the 3 steps under 4 different SNRs, $-3, 3, 9$, and $15$. 
The reward, SSIM score, adversarial error rate $\text{Adv}(\boldsymbol{x}^i,\boldsymbol{x}^i_{adv},\boldsymbol{x}^i_{final})$, and purification error rate $\text{Err}(\boldsymbol{x}^i,\boldsymbol{x}^i_{adv},\boldsymbol{x}^i_{final})$ in the whole training phase are shown in Fig. 4.

As illustrated in Fig. 4 (a), the reward of the model is the average reward of the 4 SNR conditions, after 2,500 steps of random action it converges after 5,000 steps.
In Fig. 4 (b), the SSIM score of conditions in 3 dB, 9 dB, and 15 dB are very close and are around 0.91, while for -3 dB, the SSIM score is relatively low, around 0.78. 

The trade-off is especially distinct in Fig. 4 (c), where the model has to reach a better image reconstruction quality, which is shown as the SSIM scores, while removing the adversarial noises. 
In consideration of this, the rise of the adversarial error rate of the model in SNR=-3 dB is acceptable with the SSIM rising and purification error rate dropping. 
According to our observation, the reason why the random phase shows a lower error rate for SNR=-3 dB is that the parameters of the model are randomly initialized in a position where the actions are set to their maximum value.
The distribution of the purification error rate mainly follows the transmission conditions, in Fig. 4 (d), when the SNR increases the error rate declines.

\begin{figure*}[htbp]
    \subfigure[The robust accuracy under semantic attacks versus SNR.]{
    \begin{minipage}[t]{0.5\linewidth}
        \centering
        \includegraphics[width=3.3in]{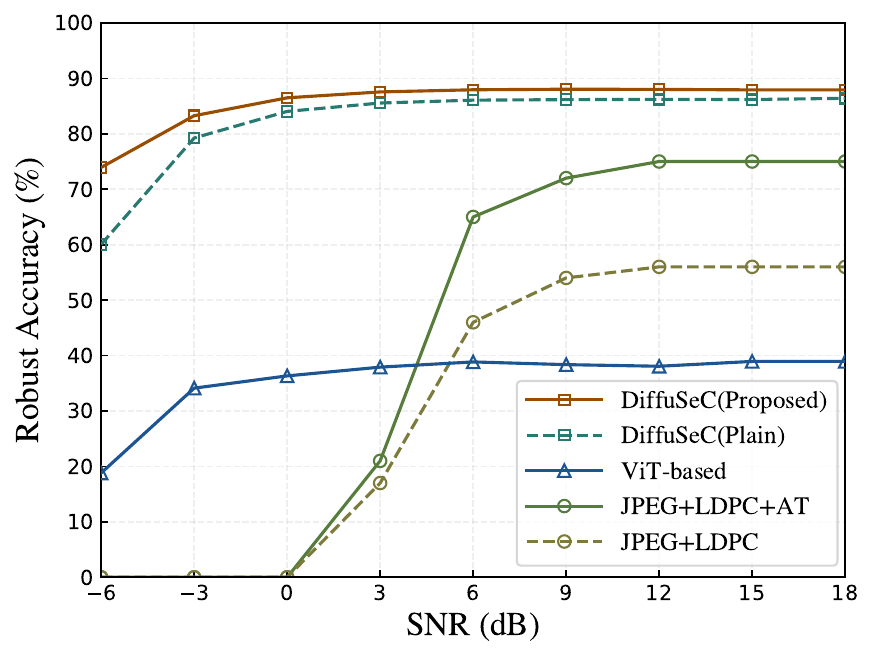}
    \end{minipage}
    }%
    \subfigure[The classification accuracy without semantic attack versus SNR.]{
    \begin{minipage}[t]{0.5\linewidth}
        \centering
        \includegraphics[width=3.3in]{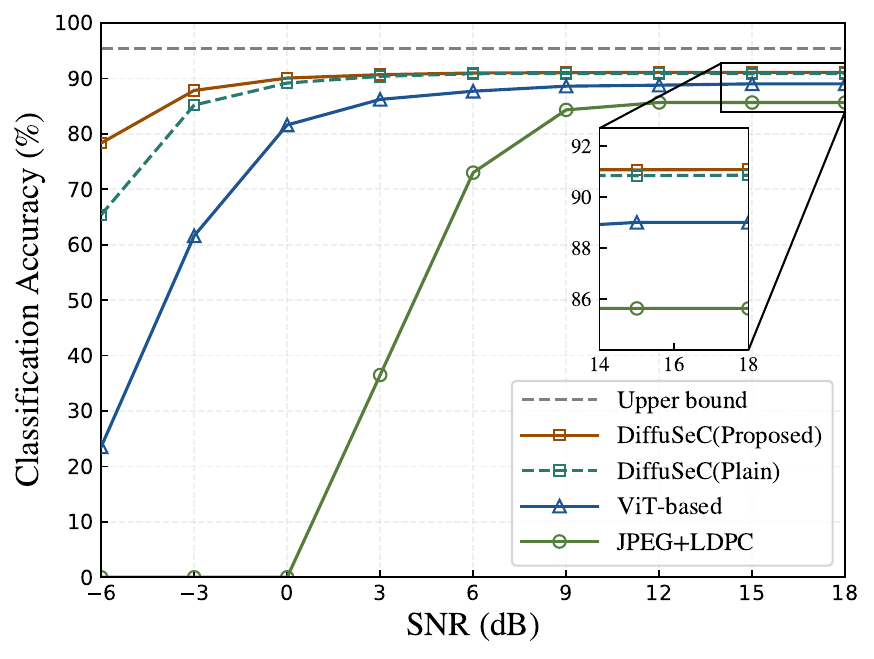}
    \end{minipage}
    }%
    \centering
    \caption{The classification accuracy and robust accuracy of the proposed DiffuSeC under different SNRs compared to conventional methods.}
\end{figure*}

\begin{table*}[!t]
\centering
\caption{The architecture of the proposed model.}
\begin{tabular}{|c|c|c|c|c|c|c|c|c|c|}
\hline
\multirow{2}{*}{Method}& \multirow{2}{*}{Standard Acc(\%)}& \multicolumn{8}{|c|}{Robust Acc(\%)}\\
\cline{3-10}
~ & ~ & SNR:-6 & SNR:-3 & SNR:0 & SNR:3 & SNR:6 & SNR:9 & SNR:12 & SNR:15\\
\hline
Masked VQ-VAE + AT\cite{hu2022robust} & 97.9 & \textbf{80.8} & 82.5 & 83.4 & 83.7 & 84.3 & 84.7 & 84.7 & 84.7\\
\hline
DiffuSeC(proposal) & 95.3 &  73.9 & \textbf{83.3} & \textbf{86.5} & \textbf{87.6} & \textbf{88.0} & \textbf{88.0} & \textbf{88.0} & \textbf{88.0} \\
\hline
\end{tabular}
\end{table*}

\subsection{System Robustness Under Semantic Attacks}
As shown in Fig. 5, we evaluate the robust accuracy and classification accuracy of the proposed system and conventional methods. 
The model of DiffuSeC is trained under the SNR from -3 dB to 12 dB, and tested under -6 dB to 18 dB.
For each SNR setting, we randomly initiate the timestep and plus step of DiffuSeC, respectively in the range of 1 to 50 and 0 to 50, and let the adaptive diffusion scheme choose the next state of the model for 3 steps to get the average robust accuracy of the model.
For DiffuSeC(plain) we adopt 20 steps for both diffusing and denoising stepsize.
In Fig. 5 (a), the robust accuracy of DiffuSeC and its plain version is compared to ViT-based semantic communication and JPEG+LDPC. 
Robust accuracy stands for the classification accuracy of a model under adversarial attacks.
When SNR is in the range of -6 dB to 6 dB, the robust accuracy of the proposed method rises, especially when the SNR is less than -3 dB. 
This phenomenon could be caused by the training SNR range of the model.
It's obvious in Fig. 5 (a) that the asymmetric diffusion scheme increases the model performance in the low SNR region with extra denoise steps, and the channel-adaptive scheme reduces the error rate by adaptively modifying the diffusion timestep based on different transmission conditions.
In Fig. 5 (b), the classification accuracy without adversarial attacks under different SNR conditions is illustrated.
Overall the classification accuracy of DiffuSeC surpassed that of its plain version, while they converge in performance when SNR is over 3 dB.
The proposed DiffuSeC reaches a top classification accuracy of 91.1\%.

As illustrated in Table II, we compare the standard accuracy and robust accuracy of our proposal and Masked VQ-VAE + AT, which is another novel architecture of robust semantic communication, under different channel states.
For our raw WRN-28-10, the standard accuracy is 95.3\%, but the robust accuracy under PGD attack is 0\%.
The robust accuracy of our model is also tested following the method we adopted in Fig. 5, in which the model is gradually optimized by the adaptive diffusion scheme in 3 steps.

We chose this evaluating method, instead of picking the final state after more steps of adjustment, because this method shows the stability of our model in fluctuating transmission environments, and how it quickly adapts to the new environment.
The condition of the physical channel can be very unstable when the channel attack is implemented, in this case, the model needs to change from any beginning state and adapt to the channel statement as fast as possible. 
Although our model is shown to be less robust when the SNR is at -6 dB, DiffuSeC successfully maintains higher robust accuracy than Masked VQ-VAE when the SNR is larger than 3 dB, even with a classification model with lower standard accuracy.

We give some examples of the original clean images, the images after their data sourse attacks, and the images received before and after the asymmetric denoising module in SNR=-3 dB, SNR=3 dB, SNR=9 dB, and SNR=15 dB in Fig. 6.
The better transmission condition lets the images reach high scores, but when in low SNRs, the DiffuSeC is still able to maintain the image quality and details.

\begin{figure*}[tbp]
\centering
\includegraphics[width=0.9\linewidth]{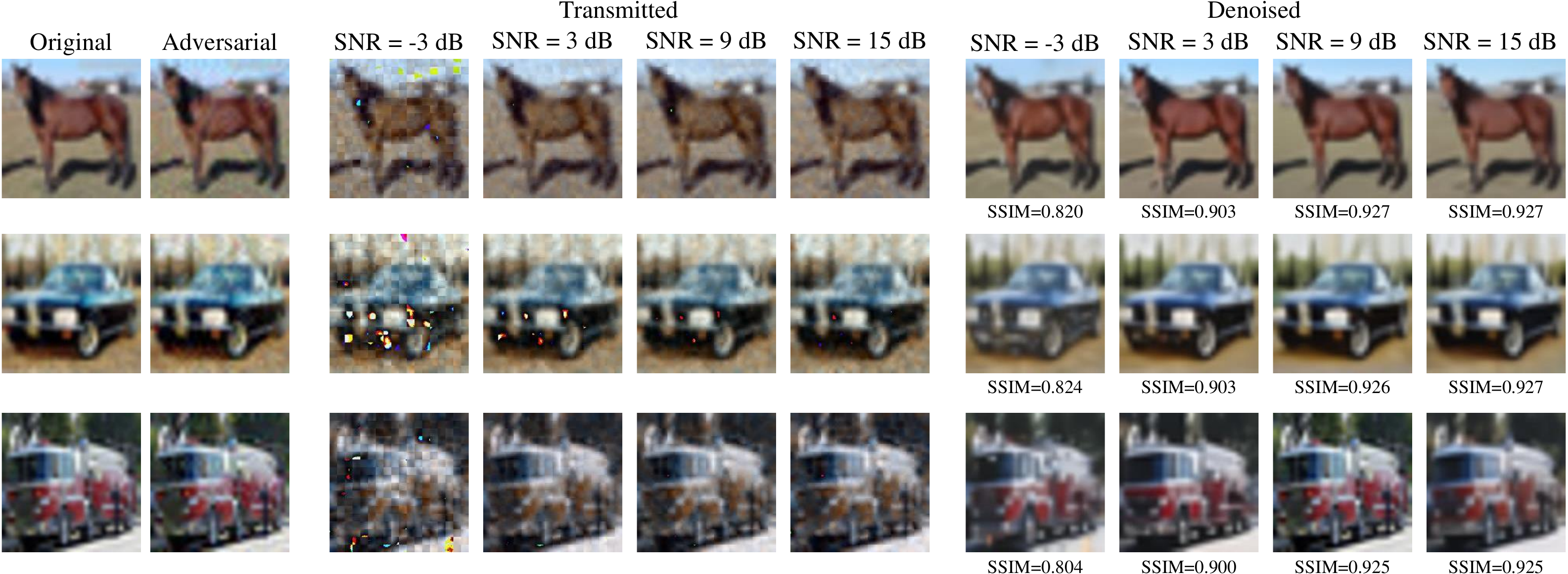}
\caption{The original images with their adversarial versions and the transmitted images and final results after going through the DiffuSeC.}
\end{figure*}

\section{Conclusion}

In this paper, we have addressed the critical issue of semantic attacks in semantic communication systems by introducing DiffuSeC, a secure semantic communication system that leverages the diffusion model and DRL to mitigate vulnerabilities caused by semantic attacks. 
For the data source attacks and channel attacks, we developed two modules named the diffusing module and the asymmetric denoising module to mitigate the impact of semantic attacks. 
In particular, we introduced a DDPG-based channel-adaptive diffusion step selection scheme, which supports the step selection of the two modules, and improves the system's robustness under fluctuating transmission conditions caused by channel semantic attacks. 
Our simulation results demonstrate that DiffuSeC outperforms previous works with impressive robust accuracies under most SNR regions, and reaches a top robust accuracy of 88.0\% and top classification accuracy of 91.1\%. 
Moreover, it exhibits the capability to adapt rapidly to varying SNRs in unstable environments. 
This research advances the state of semantic communication security and contributes a practical solution to enhance its robustness.

\bibliographystyle{IEEEtran}
\bibliography{ref}

\begin{IEEEbiography}[{\includegraphics[width=1in,height=1.25in,clip,keepaspectratio]{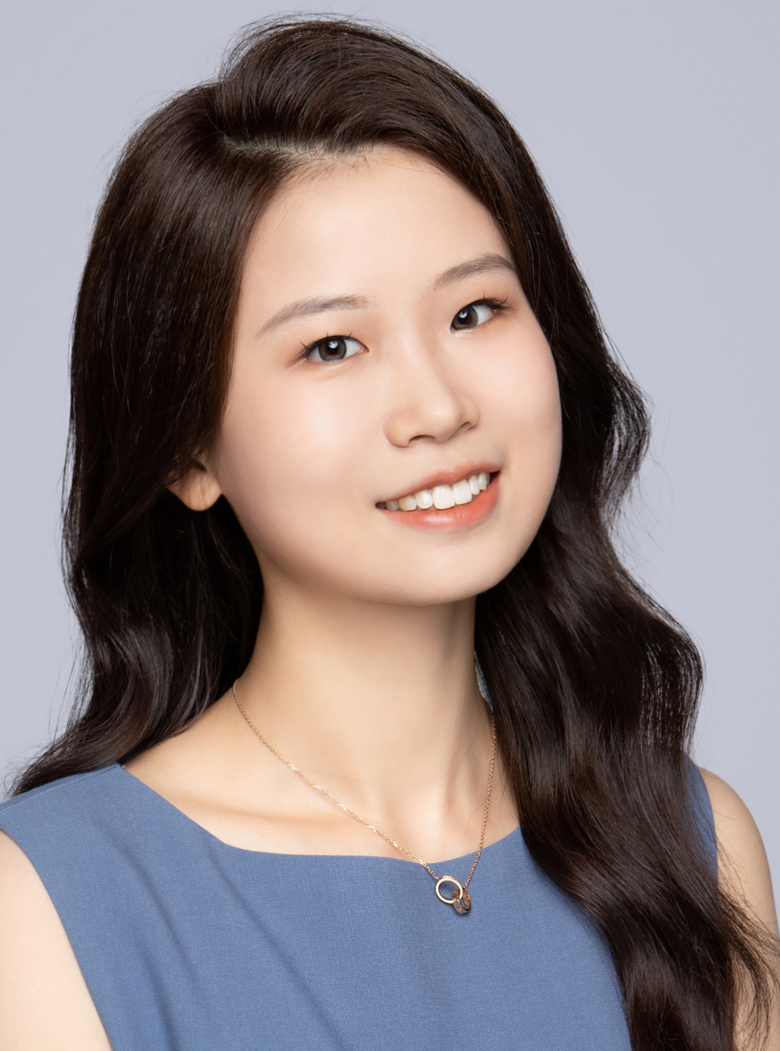}}]{Xintian Ren}
(Student Member, IEEE) received the B.S. degree in cyber science and engineering from Southeast University, Nanjing, China. She is currently pursuing her M.S. degree in information security engineering at the School of Electronic Information and Electrical Engineering, Shanghai Jiao Tong University, Shanghai, China. Her research interest includes semantic communication and machine learning. 
\end{IEEEbiography}

\vspace{-7pt}

\begin{IEEEbiography}[{\includegraphics[width=1in,height=1.25in,clip,keepaspectratio]{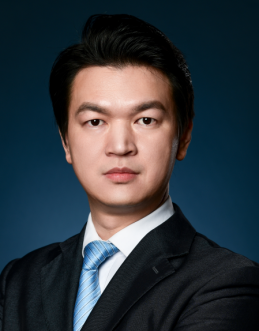}}]{Jun Wu}
(Senior Member, IEEE) received the Ph.D. degree in information and telecommunication studies from Waseda University, Japan, in 2011. He is currently a professor with the Graduate School of Information, Production and Systems of the same university. He is the chair of IEEE P21451-1-5 Standard Working Group for Internet of things. His research interests include the intelligence and security techniques of Internet of Things (IoT), edge computing, big data, 5G/6G, etc. He is the author or co-author of more than 200 peer-reviewed journal/conference papers within the above-mentioned topics. His publications have received a few distinctions, which includes the Best Paper Award of IEEE Transactions on Emerging Topics in Computing, in 2020, Best Paper Award of International Conference on Telecommunications and Signal Process in 2019, Best Conference Paper Award of the IEEE ComSoc Technical Committee on Communications Systems Integration and Modeling in 2018. He has served as the Track Chair for VTC 2019, 2020, 2023 and the TPC Member of more than ten international conferences including ICC, GLOBECOM, etc. He severs as an Associate Editor for the IEEE Systems Journal and IEEE Networking Letters. He has served as a Guest Editor for the IEEE Transactions on Industrial Informatics, IEEE Transactions on Intelligent Transportation, IEEE Sensors Journal, Sensors, Frontiers of Information Technology \& Electronic Engineering (FITEE), etc. 
\end{IEEEbiography}
\vspace{11pt}

\begin{IEEEbiography}[{\includegraphics[width=1in,height=1.25in,clip,keepaspectratio]{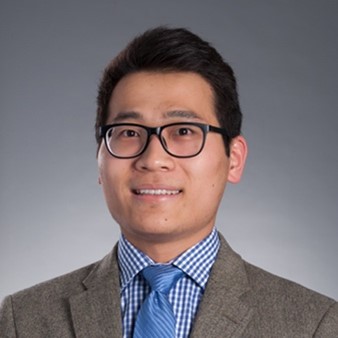}}]{Hansong Xu}
(Member, IEEE) is currently an assistant researcher of Electronic Information and Electrical Engineering, Shanghai Jiao Tong University, Shanghai, China. He was a postdoctoral researcher with Shanghai Jiao Tong University from 2020 to 2022. He obtained his Ph.D degree from the Department of Computer and Information Sciences at Towson University, MD, USA, in 2020. He received the Graduate Student Research Award and the Doctorial Research Fellowship at Towson University in 2018. He was a recipient of the Shanghai Pujiang Talent Program award and Special Support from China Postdoctoral Science Foundation. His current research interests include internet of things, machine learning, and digital twin.
\end{IEEEbiography}

\vspace{11pt}

\begin{IEEEbiography}[{\includegraphics[width=1in,height=1.25in,clip,keepaspectratio]{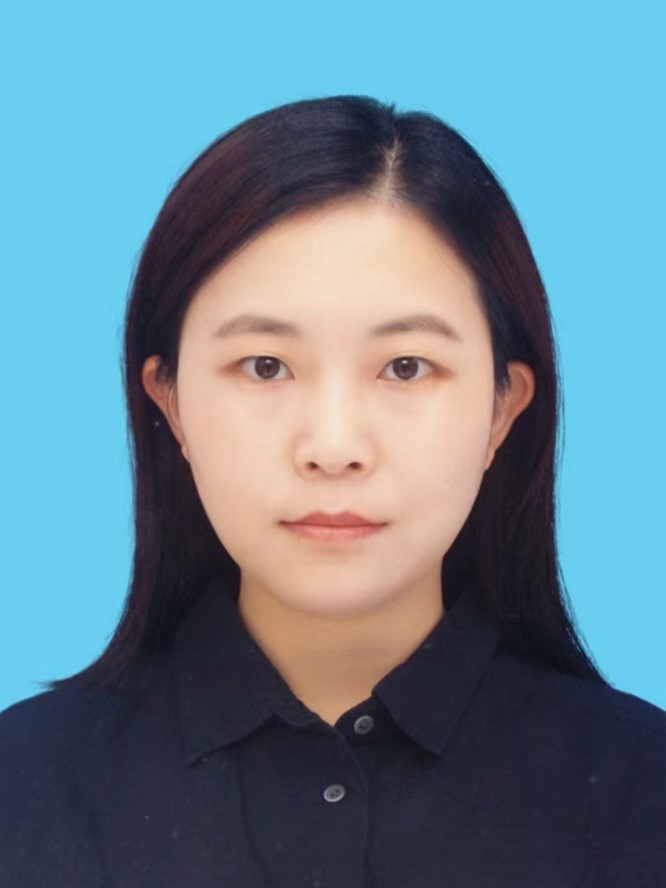}}]{Qianqian Pan}
(Member, IEEE) received B.S. and M.S. in information and communication engineering from the School of Information Science and Engineering, Southeast University, Nanjing, China, in 2015 and 2018, respectively. She received the Ph.D. degree in cyberspace security from the School of Electronic Information and Electrical Engineering, Shanghai Jiao Tong University, Shanghai, China, in 2023. From July 2022 to December 2022, she visited the Muroran Institution of Technology, Muroran, Japan. She is currently with the Graduate School of Engineering, The University of Tokyo, Tokyo, Japan. Her research interests include blockchain, privacy protection, and next-generation network security. Dr. Pan has obtained two best student paper awards of IEEE conferences. She serves as a TPC member for IEEE Vehicular Technology Conference 2023 and the reviewer for IEEE Transactions on Dependable and Secure Computing, IEEE Transactions on Industrial Informatics, etc.
    
\end{IEEEbiography}

\vfill

\end{document}